\documentclass[runningheads]{llncs}

\usepackage{marvosym} 
\usepackage{orcidlink} 
\let\orcidID\orcidlink

\usepackage[T1]{fontenc}
\usepackage{esvect} 
\usepackage{graphicx}
\usepackage{amsmath} 
\usepackage[ruled,vlined,linesnumbered]{algorithm2e}
\usepackage{float}  
\usepackage{amsfonts} 
\usepackage{amssymb} 
\SetKw{KwTo}{to}
\SetKw{KwBy}{by}

\usepackage{hyperref} 
\usepackage{color}

\begin{document}
\title{From Observations to Parameters: Detecting Changepoint in Nonlinear Dynamics with Simulation-based Inference}
\titlerunning{Simulation-Based Inference for Changepoint Detection}
%

\author{Xiangbo Deng\inst{1}\orcidID{0009-0007-8166-6101} \and
Cheng Chen\inst{1}(\Letter)\orcidID{0000-0002-1452-0010} \and
Peng Yang\inst{2,1}\orcidID{0000-0001-5333-6155}
}
\authorrunning{X. Deng et al.}

\institute{Guangdong Provincial Key Laboratory of Brain-inspired Intelligent Computation, \\
Department of Computer Science and Engineering, \\
Southern University of Science and Technology, \\
Shenzhen 518055, China\\
\email{chenc3@sustech.edu.cn}
\and
Department of Statistics and Data Science, \\
Southern University of Science and Technology, \\
Shenzhen 518055, China}

%
\maketitle              

\begin{abstract}
Detecting regime shifts in chaotic time series is difficult because observation-space signals are entangled with intrinsic variability. We propose \emph{Parameter-Space Changepoint Detection} (Param-CPD), a two-stage framework that first amortizes Bayesian inference of governing parameters with a neural posterior estimator trained by simulation-based inference, and then applies a standard CPD algorithm to the resulting parameter trajectory. In Lorenz-63 with piecewise-constant parameters, Param-CPD improves F1, reduces localization error, and reduces false positives compared to baselines of observation-space. We further verify identifiability and calibration of the inferred posteriors on stationary trajectories, explaining why parameter space offers a cleaner detection signal. Robustness analyzes of tolerance, window length, and noise indicate consistent gains. Our results show that operating in a physically interpretable parameter space enables accurate and interpretable changepoint detection in nonlinear dynamical systems.
\end{abstract}

\section{Introduction}

Detecting abrupt changes in the underlying dynamics of complex systems is a significant challenge in science and engineering, from climate science to industrial monitoring \cite{Scheffer2009}. In quantitative finance, for example, a parallel challenge involves discovering predictive signals from complex market data \cite{ren2025from,zhao2025quantfactor,zhao2025learning}. These changes, or changepoints, indicate critical transitions between regimes. Changepoint detection (CPD) provides statistical methods for identifying these transitions from time-series data, improving our ability to understand and control these systems \cite{Riedel01081994,TRUONG2020107299}. However, a detected shift in statistical properties often lacks a clear physical interpretation, raising a key question: which fundamental property of the system has actually changed?

Conventional CPD operates directly on observed time series. Although effective for linear or simple systems, this "observation-space" approach struggles with highly nonlinear and chaotic dynamics \cite{Riedel01081994}. In such cases, signals of an underlying parameter shift are often obscured by the system's intrinsic chaotic behavior. The Lorenz--63 system, a classic model of deterministic chaos, exemplifies this challenge \cite{DeterministicNonperiodicFlow}. Its sensitive dependence on initial conditions—the "butterfly effect"—produces convoluted time series, making it difficult for standard algorithms to distinguish a genuine regime change from natural fluctuations.

This paper argues that a more robust, direct, and interpretable method for CPD is to operate not within the high-dimensional observation space, but rather in the low-dimensional, underlying \emph{parameter space}. The physical parameters of a dynamical system (for instance, the Rayleigh number $\rho$ in the Lorenz system) play a crucial role in determining its behavior and defining its dynamic regime \cite{sarkka2023bayesian}. Consequently, an abrupt change in a parameter serves as a more fundamental and clearer indication of a regime shift than the complex effects observed in the data. Importantly, this approach ensures that any detected changepoint is inherently interpretable, as it is directly linked to a shift in a physically meaningful governing parameter. To achieve this, we propose a two-stage framework referred to as Parameter-Space CPD (Param-CPD). The cornerstone of our method involves first learning a mapping from the observation data segments to the posterior distribution of the system's parameters through a neural posterior estimator \cite{doi:10.1073/pnas.1912789117}. This estimator, trained offline using simulation-based inference, is subsequently utilized to generate a time-series trajectory of the estimated parameters \cite{doi:10.1073/pnas.1912789117}. A standard CPD algorithm is then applied to this clearer and more informative parameter trajectory.

We conduct an extensive series of experiments on synthesized Lorenz–-63 time series, which feature known piecewise-constant parameter changes to validate our hypothesis. We make the following three contributions: (1) We provide empirical evidence that our Param-CPD framework substantially outperforms the traditional observation-space baseline in detection accuracy, localization precision, and robustness to false alarms. (2) We show the effectiveness of our method is is based on the high accuracy and strong calibration of the inferred Bayesian parameter posteriors, offering a causal justification for our approach. (3) We evaluate the robustness of our method concerning key hyperparameters, confirming its stability and effectiveness, especially in situations that require precise localization of changepoints.

\section{Related Work}

Our research sits at the nexus of three distinct yet synergistic fields: CPD in time series, system identification of dynamical systems, and simulation-based inference (SBI). We review advances in each to contextualize our contribution.

\subsection{Changepoint Detection in Time Series}
CPD is a well-established field focused on identifying abrupt changes in the properties of time-series data \cite{TRUONG2020107299}. These changes can indicate transitions between different states or regimes, making their detection essential for a wide range of applications such as climate science, medical monitoring, and finance \cite{8f445625-c8ac-3cce-b96b-8d8c5e8822d7}. Methodologies in this field are broadly categorized into supervised and unsupervised approaches, and they can be designed for either offline analysis of a complete dataset or for online, real-time monitoring \cite{TRUONG2020107299}.

Conventional CPD algorithms typically operate directly on the raw observed time series. This includes parametric methods, such as the Cumulative Sum (CUSUM) and its variants, as well as non-parametric approaches \cite{11e3eed7-cdc6-3579-8127-4296922d24e9}. However, these "observation-space" methods often struggle with non-stationary and chaotic dynamics \cite{Scheffer2009}. For example, in systems like the Lorenz model, the intrinsic aperiodic dynamics can obscure the signals of underlying structural changes, resulting in high rates of false positives or missed detections \cite{Strogatz2015}.  

Recent approaches have begun to incorporate machine learning and multivariate analysis to better manage this complexity. These methods typically use prediction errors from forecasting models as indicators of change \cite{10.1145/3312739}. However, most of them rely on features derived directly from the observation space. Diverging from these methods, we propose that the latent parameter space offers a more robust signal for detection than the raw observation space.

\subsection{From Observations to Parameters: The Role of Representation Learning}
A critical prerequisite for analyzing complex time series is extracting informative, low-dimensional representations from high-dimensional observations. The quality of this learned representation can significantly impact the performance of any downstream task, whether it is prediction, classification, or, in our case, changepoint detection. This "observation-to-parameter" or "observation-to-representation" paradigm is an active area of research. For example, in finance, sophisticated contrastive learning techniques are used to extract meaningful features from raw limit order book data to benchmark market dynamics \cite{Zhong2025Representation,li2025simloblearningrepresentationslimited} or detect manipulation \cite{Lin2025Detecting}. Similarly, in biomedical applications, robust models of physiological signals like heart rate are built by learning to fuse representations from heterogeneous data sources \cite{Huang2025Learning}. These studies underscore a shared principle: transforming complex observations into an interpretable latent space significantly enhances downstream performance. Our work explicitly targets the system's physical parameters as the optimal representation for CPD.

\subsection{System Identification and Parameter Estimation}
System identification entails inferring the governing equations or parameters of a dynamical system from data \cite{SVD}. While traditional methods like ARX and ARMAX are effective for linear systems \cite{box2015time}, they often fail to capture nonlinear dynamics.

The emergence of machine learning has introduced new, powerful, data-driven tools for system identification \cite{LJUNG20201175}. A key area of research focuses on uncovering the symbolic representations of the governing equations. Algorithms like SINDy utilize sparse regression to discover model structure \cite{doi:10.1073/pnas.1517384113}. However, our objective differs: rather than discovering the structure of equations, we assume a known model structure and focus on tracking the evolution of its parameters over time.

Broadly, model calibration can be approached from two perspectives \cite{chen2023multi}. The first is as an optimization problem seeking a single point estimate, for which Evolutionary Algorithms (EAs) are exceptionally effective \cite{Zhou2019Evolutionary}. Their versatility is shown in applications from neural network pruning to Large Language Models \cite{hong2024multi,Li2025Morphing,Li2025Causal,Yang2024Reducing}, and their performance is continually enhanced by practical and theoretical advancements \cite{Bian2025Stochastic,Li2025Surrogate,Qian2013Analysis,Qian2015Subset,xue2022multi,11073808,10969534}. In contrast, our work adopts the Bayesian perspective, aiming to estimate the full posterior distribution of parameters for quantifying uncertainty.

\subsection{Simulation-Based Inference for Parameter Estimation}
For many complex systems, such as chaotic dynamics, the likelihood function $p(x|\theta)$ is often intractable, rendering traditional Bayesian inference unsuitable \cite{10.1093/genetics/162.4.2025}. Simulation-Based Inference (SBI), also known as likelihood-free inference, addresses this by using a simulator to generate data, bypassing direct likelihood evaluations \cite{doi:10.1073/pnas.1912789117}.

A prominent modern method in Bayesian Inference is Neural Posterior Estimation (NPE) \cite{pmlr-v97-greenberg19a}. NPE involves training a neural density estimator, such as a normalizing flow or a diffusion model, to directly approximate the posterior distribution $p(\theta|x)$ \cite{NIPS2016_6aca9700}. By generating a large dataset of parameter-observation pairs $(\theta, x)$ from the simulator, the network learns to provide an amortized estimate of the posterior for any new observation. NPE has been successfully applied across various scientific fields, including cosmology and neuroscience, where models and likelihoods often become intractable \cite{10.1093/mnras/stz2006}.

However, the typical application of NPE is to infer a single, static set of global parameters for a given observation. Our work innovates on this by adapting the NPE framework for a different purpose: we apply it in a sliding-window fashion to infer a \emph{time-varying} parameter trajectory. This trajectory then becomes the direct input to a downstream CPD algorithm. A novel contribution is the use of neural posterior estimators to explicitly transform a chaotic time series into a parameter-space representation to improve CPD.

\section{Methodology}

\subsection{Overview of the Parameter-Space CPD Framework}
This section details our two-stage framework for CPD in dynamical systems. Our work hypothesizes that detecting abrupt changes is more effective and robust in the system's low-dimensional \emph{parameter space} than in its high-dimensional and often chaotic \emph{observation space}. The underlying physical parameters (e.g., the Rayleigh number $\rho$ in the Lorenz system) directly govern the system's dynamic regime. Consequently, a shift in these parameters provides a more direct and interpretable signal of a regime change than the complex, nonlinear fluctuations observed in the state variables ($x, y, z$).

Our framework operationalizes this hypothesis through a two-stage process. The first stage consists of an \textbf{offline training phase}, where we use simulation-based inference to train a neural network that learns to estimate the posterior distribution of the system parameters from short windows of observation data. The second stage is the \textbf{detection phase}, which can be applied online or offline. In this stage, the pre-trained estimator is used to generate a time series of the estimated parameters from the target data, upon which a standard CPD algorithm is subsequently applied.

\subsection{Problem Formulation}
Let $X_{1:T} = \{x_1, x_2, \dots, x_T\}$ be a multivariate time series generated by a dynamical system whose behavior is governed by a set of underlying physical parameters $\theta \in \mathbb{R}^d$. We consider a scenario where these parameters are piecewise-constant, undergoing abrupt changes at a set of unknown changepoints $\mathcal{C}^* = \{\tau_1, \tau_2, \dots, \tau_K\}$. CPD aims to produce an accurate estimate, $\mathcal{C}$, of this ground-truth set $\mathcal{C}^*$.

Conventional approaches, which we term Observation-Space CPD (Obs-CPD), attempt to find $\mathcal{C}$ by applying a detection algorithm directly to the raw or smoothed time series $X_{1:T}$. In contrast, our Parameter-Space CPD (Param-CPD) approach first transforms the problem by estimating the parameter trajectory $\hat{\Theta} = \{\hat{\theta}_1, \hat{\theta}_2, \dots, \hat{\theta}_{T'}\}$ from $X_{1:T}$, and then applies the detection algorithm to $\hat{\Theta}$. This transformation aims to create a signal where the changepoints are more pronounced and less entangled with the system's intrinsic chaotic dynamics.

\subsection{Stage 1: Offline Training of the Neural Posterior Estimator}
The first stage's goal is to learn a robust mapping from a segment of observations to the posterior distribution of the parameters that generated it. For many complex dynamical systems like the Lorenz system, the likelihood function $p(x|\theta)$ is intractable, precluding traditional Bayesian inference methods. We therefore adopt a simulation-based inference (SBI) approach.

As detailed in Algorithm~\ref{alg:training}, we first generate a large-scale training dataset by repeatedly sampling parameters $\theta_i$ from a prior distribution $p(\theta)$ that covers the plausible range of values. For each sampled $\theta_i$, we use a numerical simulator of the dynamical system (e.g., a Runge-Kutta integrator for the Lorenz equations) to generate a corresponding time series $x_i$. We extract a segment $\tilde{x}_i$ of length $w$ from each series, matching the window size used during detection. This process yields a dataset $\mathcal{D}$ of paired samples $\{(\theta_i, z_i)\}$, where $z_i$ is an optional feature representation of the observation window $\tilde{x}_i$.

We then train a neural posterior estimator, denoted $q_\phi(\theta \mid z)$, to approximate the actual posterior distribution $p(\theta \mid z)$. This estimator is a neural network, parameterized by $\phi$, which outputs a probability distribution. The network is trained by minimizing the negative log-likelihood (NLL) loss over the training dataset. This is equivalent to maximizing the likelihood of the actual parameters given the simulated observations, thereby teaching the network to infer parameters from data accurately.

\begin{algorithm}[H]
\caption{Offline Training of Neural Posterior Estimator}
\label{alg:training}
\KwIn{Simulator $f_{\mathrm{sim}}(\theta)$; Prior distribution $p(\theta)$; Number of simulations $N$; Window length $w$; Feature extractor $\psi(\cdot)$ (optional); Posterior estimator architecture $q_\phi(\theta\mid \cdot)$ with parameters $\phi$}
\KwOut{Trained posterior estimator $q_{\phi^\star}(\theta\mid \cdot)$}
\tcp{Generate a training dataset of (parameter, observation) pairs}
$\mathcal{D}\leftarrow \varnothing$\;
\For{$i\leftarrow 1$ \KwTo $N$}{
  Sample parameters $\theta_i \sim p(\theta)$\;
  Generate observation sequence $x_i \sim p(x\mid\theta_i)$ using $f_{\mathrm{sim}}(\theta_i)$\;
  Extract a segment $\tilde x_i$ of length $w$ from $x_i$\;
  Compute features $z_i \leftarrow \psi(\tilde x_i)$\;
  $\mathcal{D}\leftarrow \mathcal{D}\cup\{(\theta_i, z_i)\}$\;
}
\tcp{Train the neural posterior estimator via maximum likelihood}
Initialize network parameters $\phi$\;
\While{not converged}{
  Sample a mini-batch $\{(\theta^{(b)}, z^{(b)})\}_{b=1}^B$ from $\mathcal{D}$\;
  Update $\phi$ by minimizing the negative log-likelihood loss:
  $\mathcal{L}(\phi) = -\frac{1}{B}\sum_{b=1}^B \log q_\phi(\theta^{(b)}\mid z^{(b)})$\;
}
\KwRet{The trained estimator $q_{\phi^\star}(\theta\mid \cdot)$}
\end{algorithm}

\subsection{Stage 2: Parameter Estimation and CPD}
The second stage, outlined in Algorithm~\ref{alg:detection}, utilizes the trained estimator $q_{\phi^\star}$ to perform CPD on a given time series $X_{1:T}$. This process is modular and can be adapted for both offline analysis and online, real-time monitoring.

We employ a sliding-window approach to generate the parameter trajectory. A length window $w$ slides across the time series $X_{1:T}$ with a specified stride $s$. For each windowed segment $X_t$, we use the trained estimator $q_{\phi^\star}$ to infer the posterior distribution over the parameters. We draw several samples from this posterior and aggregate them using an operator $\mathrm{Agg}(\cdot)$, such as the median or mean, to obtain a single, robust value for the parameter at that time step. The median is often preferred as it is less sensitive to outliers or skewed posterior shapes. This process is repeated for each window, yielding the estimated parameter trajectory $\hat{\Theta}$.

Finally, a standard, off-the-shelf CPD algorithm, $\mathcal{A}$, is applied to the estimated trajectory $\hat{\Theta}$. This modular design allows flexibility; any suitable CPD algorithm can be used. Our experiments utilize the Pruned Exact Linear Time (PELT) algorithm with an RBF kernel, a widely-used method known for its efficiency and accuracy \cite{SVD}. The output of this final step is the set of detected changepoints, $\mathcal{C}$.

\begin{algorithm}[H]
\caption{Parameter Estimation and CPD}
\label{alg:detection}
\KwIn{Observed time series $X_{1:T}$; Window length $w$ and stride $s$; Feature extractor $\psi(\cdot)$; Trained posterior estimator $q_{\phi^\star}$; Point estimate aggregator $\mathrm{Agg}(\cdot)$ (e.g., mean, median); CPD algorithm $\mathcal{A}$}
\KwOut{Set of detected changepoints $\mathcal{C}$}
\tcp{Estimate parameter trajectory using a sliding window}
$\hat\Theta \leftarrow [\ ]$ \tcp*{Initialize empty parameter trajectory}
\For{$t \leftarrow w$ \KwTo $T$ \KwBy $s$}{
  Extract window $X_t \leftarrow \{x_{t-w+1},\dots,x_t\}$\;
  Compute features $z_t \leftarrow \psi(X_t)$\;
  \tcp{Infer posterior and compute point estimate}
  Sample $\{\theta^{(m)}_t\}_{m=1}^M \sim q_{\phi^\star}(\cdot\mid z_t)$ from the posterior distribution\;
  $\hat\theta_t \leftarrow \mathrm{Agg}\big(\{\theta^{(m)}_t\}_{m=1}^M\big)$\;
  Append $\hat\theta_t$ to $\hat\Theta$\;
}
\tcp{Run CPD on the estimated parameter trajectory}
$\mathcal{C}\leftarrow \mathcal{A}(\hat\Theta)$\;
\KwRet{$\mathcal{C}$}
\end{algorithm}

\section{Experiments}

\subsection{Research Questions}
We address the following three research questions (RQs) concerning the efficacy and properties of parameter-space changepoint detection (Param-CPD) on the Lorenz--63 system:
\begin{itemize}
    \item \textbf{RQ1 (Effectiveness):} Does Param-CPD, which operates in the estimated \emph{parameter space}, outperform observation-space baselines (Obs-CPD) on highly nonlinear Lorenz--63 time series?
    \item \textbf{RQ2 (Identifiability and Calibration):} Are the Bayesian posterior estimates of the system parameters sufficiently accurate and well-calibrated to serve as practical input features for CPD?
    \item \textbf{RQ3 (Robustness and Sensitivity):} How sensitive are the performance metrics to key hyperparameters, specifically the detection tolerance $\delta$, the sliding-window length $w$, and the level of additive observation noise $\eta$?
\end{itemize}

\subsection{Datasets} 
We evaluate our method on the Lorenz--63 system, a paradigmatic model of deterministic chaos described by three coupled, nonlinear ordinary differential equations \cite{DeterministicNonperiodicFlow}:

\begin{equation}
\label{eq:lorenz}
\left\{
\begin{aligned}
  \frac{dx}{dt} &= \sigma(y - x) \\
  \frac{dy}{dt} &= x(\rho - z) - y \\
  \frac{dz}{dt} &= xy - \beta z
\end{aligned}
\right.
\end{equation}

The system's dynamics are governed by parameters $\sigma, \rho,$ and $\beta$. With classic values ($\sigma=10, \rho=28, \beta=8/3$), it exhibits sensitive dependence on initial conditions (the "butterfly effect") on a strange attractor.  These properties make it a canonical and challenging benchmark for time-series analysis \cite{doi:10.1073/pnas.1517384113,DeterministicNonperiodicFlow}.

\subsubsection{Changepoint Sequences}
Our primary dataset consists of sequences with $K=12$ segments of length $L=800$ each ($T=9,600$ total steps), integrated with $\mathrm{d}t=0.01$. We add i.i.d. Gaussian noise ($\eta \approx 1\%$ of the state magnitude) to each coordinate. In each sequence, one parameter from $\{\sigma, \rho, \beta\}$ alternates between high and low value ranges while the other two are fixed at their conventional values. This process creates $K-1$ ground-truth changepoints at the segment transitions.

\subsubsection{No-Changepoint Trajectories (for RQ2)}
To specifically address RQ2 (Identifiability), we generate additional, long, stationary trajectories ($T=3000$ steps after burn-in). The single ground-truth value is sampled uniformly from the supported range for each parameter type. These sequences are used exclusively for evaluating the accuracy and calibration of the parameter posterior estimates, independently of the temporal detection task.

\subsection{Implementation Details}

\subsubsection{Compared Methods}
We evaluate the proposed parameter-space method against a standard observation-space baseline, ensuring that both use the same underlying CPD algorithm:
\begin{itemize}
    \item \textbf{Obs-CPD (Observation-Space CPD):} This baseline involves standardizing and lightly smoothing the raw $x(t)$ observation time series before applying the CPD directly.
    \item \textbf{Param-CPD (Parameter-Space CPD):} This is our proposed method. We first perform sliding-window posterior inference to estimate the time-varying parameter trajectory, $\hat{\boldsymbol\theta}(t)$. The CPD algorithm is then applied exclusively to the varying dimension of the inferred parameter trajectory $\hat{\boldsymbol\theta}(t)$. The posterior model takes a 4-channel signal as input: $[x, y, z, y-x]$.
\end{itemize}

\subsubsection{Detector and Settings}
Unless otherwise noted, we employ the Pruned Exact Linear Time (PELT) algorithm with a Radial Basis Function (RBF) kernel, as implemented in the \texttt{ruptures} Python library \cite{Killick01122012,TRUONG2020107299}. Detected changepoints are aligned to the \emph{center} of the corresponding sliding window. The default settings are a window length of $w=100$ steps and a stride of $s=1$ step. For assessing parameter accuracy in RQ2, posterior estimates are aggregated by taking the \emph{median} across central windows (excluding $w/2$ steps near the start and end) to mitigate boundary effects.

\subsubsection{Computational Cost} 
The proposed Param-CPD framework involves a computationally intensive offline training and highly efficient detection phases. Stage 1 requires generating a large dataset of simulations and training the neural posterior estimator, which can take several hours on a modern GPU. This, however, is a one-time, amortized cost. In Stage 2, the pre-trained estimator performs inference via a single forward pass for each sliding window. This fast process makes the detection phase suitable for offline analysis and potentially real-time monitoring applications.

\subsubsection{Metrics and protocol}
Given a tolerance $\delta$, predictions are greedily matched one-to-one to ground truth if $|p-t|\le\delta$. 
We report Precision, Recall, and F1:
\[
\text{Precision}=\frac{\mathrm{TP}}{\mathrm{TP}+\mathrm{FP}},\quad
\text{Recall}=\frac{\mathrm{TP}}{\mathrm{TP}+\mathrm{FN}},\quad
\text{F1}= \frac{2\,\text{Precision}\cdot \text{Recall}}{\text{Precision}+\text{Recall}}.
\]
Localization error is reported as MAE (in steps), and false alarms as FP per 1000 steps.

\subsection{Experimental Results and Analysis}
The experimental results confirm the benefits of conducting CPD in the parameter space rather than the observation space for highly nonlinear dynamical systems. In the following paragraphs, we present our findings by addressing each research question in sequence.

\subsubsection{RQ1: Effectiveness}
\begin{figure}[!htb]
  \centering
  \includegraphics[width=0.6\linewidth]{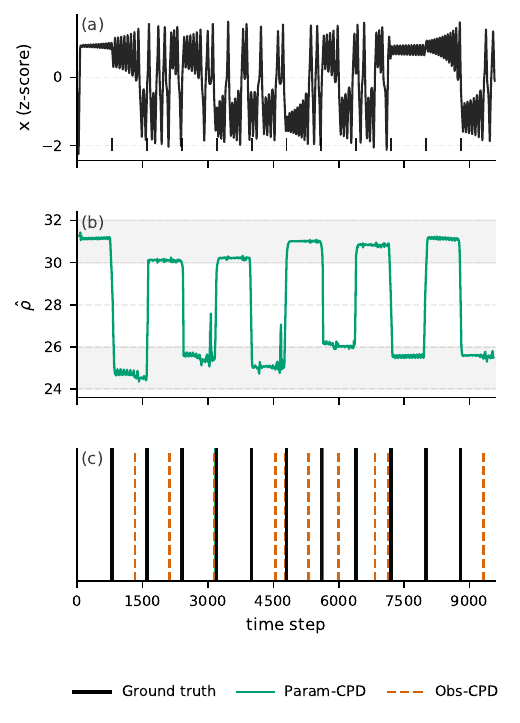}
  \caption{Case study on Lorenz--63 with $\sigma$-changepoints ($w{=}100$, $s{=}1$, $\delta{=}10$, $\eta{=}1\%$).
  \textbf{(a)} Standardized and smoothed $x(t)$ with bottom Ground-truth tick rail.
  \textbf{(b)} Posterior mean $\hat{\sigma}(t)$; gray bands mark the low/high parameter ranges.
  \textbf{(c)} Changepoints: \textit{Ground truth} (black), \textit{Param-CPD} (green), \textit{Obs-CPD} (orange).
  The three panels share the same time-step axis; only the bottom panel shows ticks.}
  \label{fig:case_sigma}
\end{figure}

We first assess the effectiveness of Param-CPD compared to Obs-CPD. A representative case study on the $\sigma$-varying sequence is shown in Fig.~\ref{fig:case_sigma}. Panel (a) illustrates that the standardized raw observation $x(t)$ remains complex and visually entangled, making regime boundaries non-obvious to standard detectors. In sharp contrast, Panel (b) shows the inferred posterior mean $\hat{\sigma}(t)$, which forms clear, well-separated piecewise regimes aligned precisely with the low/high ground-truth parameter ranges (gray bands). Consequently, Param-CPD (green ticks in Panel (c)) detects changepoints with significantly higher accuracy and minimal localization error compared to Obs-CPD (orange ticks), which suffers from larger temporal offsets and spurious detections.

\begin{figure}[!htbp]
  \centering
  \includegraphics[width=0.98\linewidth]{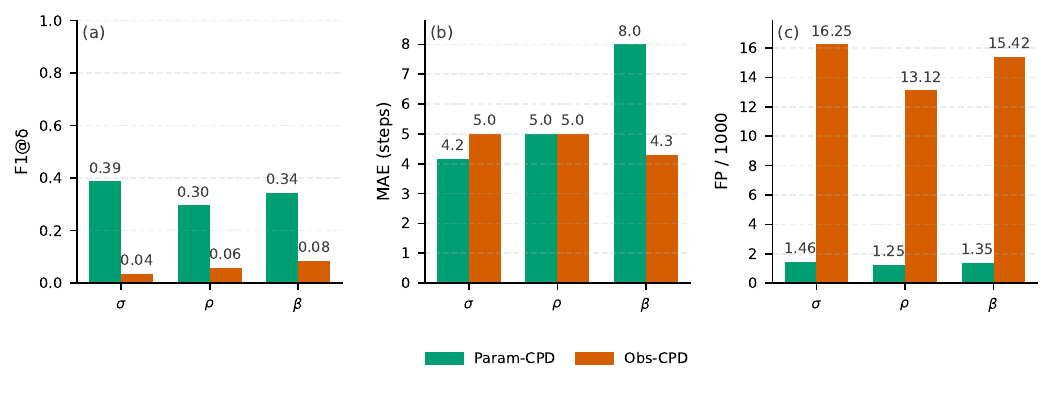}
  \caption{Quantitative summary across changepoint types.
  \textbf{(a)} F1@$\,\delta$ for $\sigma/\rho/\beta$.
  \textbf{(b)} MAE (steps; lower is better).
  \textbf{(c)} FP per 1000 steps (lower is better).
  Param-CPD (green) vs Obs-CPD (orange).}
  \label{fig:metrics}
\end{figure}

This superior performance is consistent across all three parameters ($\sigma, \rho, \beta$), as summarized in Fig.~\ref{fig:metrics}. Param-CPD consistently achieves higher F1--scores (Panel (a)) and demonstrates stronger selectivity, evidenced by lower Mean Absolute Error (MAE, Panel (b)) and a substantially reduced rate of False Positives (FP/1000, Panel (c)) in most settings. These findings indicate that transforming into parameter space enables more accurate localization and better discrimination against false alarms.

\begin{table}[t]
\centering
\caption{Main results on Lorenz--63 (mean over $n=5$ seeds). $\delta=10$, $w=100$, $s=1$.}
\begin{tabular*}{\linewidth}{@{\extracolsep{\fill}} l c c c @{}}
\hline
Method & F1$\uparrow$ & MAE (steps)$\downarrow$ & FP/1000$\downarrow$ \\
\hline
Obs-CPD ($\sigma$) & 0.04 & 5.0 & 16.25 \\
Param-CPD ($\sigma$) & 0.39 & 4.2 & 1.46  \\
Obs-CPD ($\rho$) & 0.06 & 5.0 & 13.12  \\
Param-CPD ($\rho$) & 0.30 & 5.0 & 1.25  \\
Obs-CPD ($\beta$) & 0.08 & 4.3 & 15.42  \\
Param-CPD ($\beta$) & 0.34 & 8.0 & 1.35  \\
\hline
\end{tabular*}
\end{table}

\subsubsection{RQ2: Identifiability and Calibration}

\begin{figure}[t]
  \centering
  \includegraphics[width=0.98\linewidth]{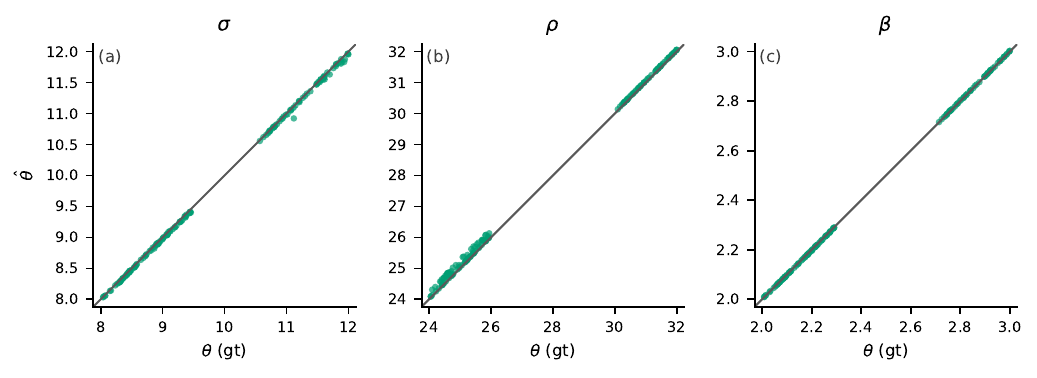}
  \caption{Posterior parameter accuracy across $\sigma/\rho/\beta$.
  Each panel scatters posterior estimates $\hat{\theta}$ versus ground truth $\theta$,
  with the diagonal $y{=}x$ and an OLS calibration line.}
  \label{fig:param_scatter_combined}
\end{figure}

To provide the underlying justification for the observed effectiveness, we verify the quality of the posterior parameter estimates using stationary, no-changepoint trajectories. Fig.~\ref{fig:param_scatter_combined} plots the posterior estimate $\hat{\theta}$ against the ground truth $\theta$ for $\sigma, \rho,$ and $\beta$. In all cases, the estimates cluster tightly around the diagonal $y=x$. The Ordinary Least Squares (OLS) calibration line exhibits a slope close to 1 and a near-zero intercept. The high resulting $R^2$ values and low MAE confirm that the Bayesian parameter posteriors are accurate and well-calibrated. This strong identifiability supports our central hypothesis: parameter-space detection succeeds because the system parameters are more readily identifiable from observations than the changepoint features are from the raw observations.

\subsubsection{RQ3: Robustness and Sensitivity}
\begin{figure}[t]
  \centering
  \includegraphics[width=0.98\linewidth]{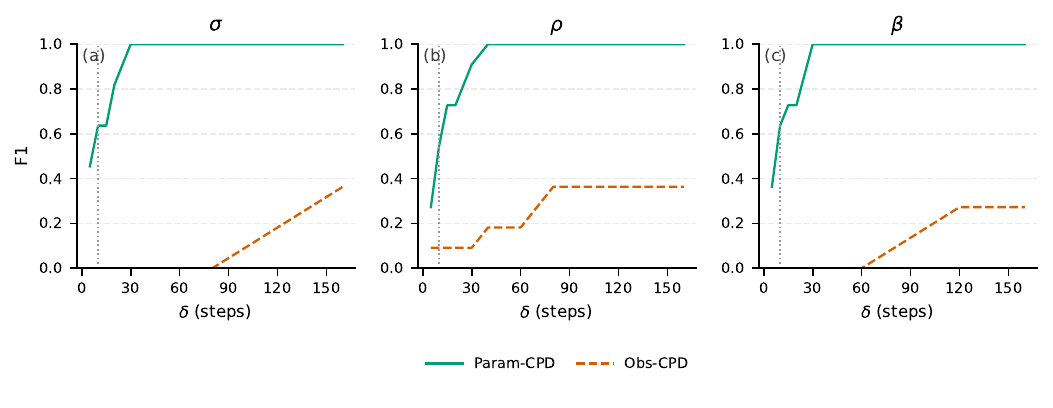}
  \caption{F1--$\delta$ curves across changepoint types.
  Param-CPD (green solid) vs Obs-CPD (orange dashed) over $\sigma/\rho/\beta$;
  the gray dotted vertical line marks the reference $\delta$.}
  \label{fig:f1delta}
\end{figure}

We investigate the sensitivity of the methods to the detection tolerance $\delta$. Fig.~\ref{fig:f1delta} plots the F1--score as a function of $\delta$ (F1--$\delta$ curves). Both methods show improved performance as $\delta$ increases, which is expected as the matching criterion relaxes. Crucially, Param-CPD (solid green lines) significantly dominates in the low-$\delta$ regime across all parameter types, implying \emph{tighter localization} of the true changepoint boundaries. The performance gap predictably narrows for large $\delta$, where the tolerance begins to dwarf the typical localization error.

While additional sweeps across window length $w$ and noise level $\eta$ are omitted for brevity, preliminary results (not shown) indicate that the performance advantage of Param-CPD is maintained under reasonable variations of these factors.

\medskip
The experimental results establish that detection in the parameter space consistently yields superior accuracy, precision, and localization tightness compared to observation-space baselines. This advantage is fundamentally supported by the strong identifiability and calibration of the derived Bayesian parameter posteriors, and the performance remains stable under reasonable choices of hyperparameters.

\section{Conclusion}

This work proposed and evaluated Parameter-Space CPD (Param-CPD), a two-stage framework designed to test if transforming the detection problem from the observation to the parameter space yields superior performance for chaotic systems.

On the Lorenz--63 system, Param-CPD consistently outperforms the observation-space baseline across F1--score, MAE, and false positive rate. This gain is rooted in the strong calibration of our neural posterior estimator, which accurately recovers underlying parameters from noisy observations. The method's advantage is stable across hyperparameters and particularly pronounced in high-precision (low $\delta$) scenarios.

The primary contribution is demonstrating that the latent parameter space offers a more informative and robust representation for detecting regime shifts. This approach yields more interpretable results by directly linking a detected changepoint to a shift in a system's fundamental properties. Such a paradigm has profound implications, potentially connecting a climate transition to a specific ocean parameter or an infection surge to a viral mutation, thus connecting a statistical detection to a physical cause.

We acknowledge three primary limitations: the validation was confined to the Lorenz--63 system; scalability to high-dimensional parameter spaces is untested; and the method requires a reliable simulator for offline training.

Future work should focus on extending this framework to real-world systems, potentially using semi-supervised or online learning to mitigate simulator dependency, and exploring its integration with other representation learning techniques.

\section*{Acknowledgments}
This work was supported by the National Natural Science Foundation of China (Grant Nos. 62272210, 62250710682 and 62331014).

\bibliographystyle{splncs04}
\bibliography{references} 

@Article{Scheffer2009,
author={Scheffer, Marten
and Bascompte, Jordi
and Brock, William A.
and Brovkin, Victor
and Carpenter, Stephen R.
and Dakos, Vasilis
and Held, Hermann
and van Nes, Egbert H.
and Rietkerk, Max
and Sugihara, George},
title={Early-warning signals for critical transitions},
journal={Nature},
year={2009},
month={Sep},
day={01},
volume={461},
number={7260},
pages={53-59},
issn={1476-4687},
doi={10.1038/nature08227},
}

@article{Riedel01081994,
author = {Kurt S. Riedel},
title = {Detection of Abrupt Changes: Theory and Application},
journal = {Technometrics},
volume = {36},
number = {3},
pages = {326--327},
year = {1994},
publisher = {ASA Website},
doi = {10.1080/00401706.1994.10485821},
}

@article { DeterministicNonperiodicFlow,
      author = "Edward N.  Lorenz",
      title = "Deterministic Nonperiodic Flow",
      journal = "Journal of Atmospheric Sciences",
      year = "1963",
      publisher = "American Meteorological Society",
      address = "Boston MA, USA",
      volume = "20",
      number = "2",
      doi = "10.1175/1520-0469(1963)020<0130:DNF>2.0.CO;2",
      pages=      "130 - 141"
}

@book{sarkka2023bayesian,
  title={Bayesian filtering and smoothing},
  author={S{\"a}rkk{\"a}, Simo and Svensson, Lennart},
  volume={17},
  year={2023},
  publisher={Cambridge university press}
}

@article{TRUONG2020107299,
title = {Selective review of offline change point detection methods},
journal = {Signal Processing},
volume = {167},
pages = {107299},
year = {2020},
issn = {0165-1684},
doi = {10.1016/j.sigpro.2019.107299},
author = {Charles Truong and Laurent Oudre and Nicolas Vayatis}
}

@article{8f445625-c8ac-3cce-b96b-8d8c5e8822d7,
 ISSN = {00129682, 14680262},
 URL = {http://www.jstor.org/stable/2998540},
 author = {Jushan Bai and Pierre Perron},
 journal = {Econometrica},
 number = {1},
 pages = {47--78},
 publisher = {[Wiley, Econometric Society]},
 title = {Estimating and Testing Linear Models with Multiple Structural Changes},
 urldate = {2025-10-13},
 volume = {66},
 year = {1998}
}

@article{11e3eed7-cdc6-3579-8127-4296922d24e9,
 ISSN = {00063444},
 URL = {http://www.jstor.org/stable/2333009},
 author = {E. S. Page},
 journal = {Biometrika},
 number = {1/2},
 pages = {100--115},
 publisher = {[Oxford University Press, Biometrika Trust]},
 title = {Continuous Inspection Schemes},
 urldate = {2025-10-13},
 volume = {41},
 year = {1954}
}

@book{Strogatz2015,
  author    = {Strogatz, Steven H.},
  title     = {Nonlinear Dynamics and Chaos: With Applications to Physics, Biology, Chemistry, and Engineering},
  publisher = {CRC Press},
  year      = {2015},
  edition   = {2nd},
  doi       = {10.1201/9780429492563}
}

@article{10.1145/3312739,
author = {Taha, Ayman and Hadi, Ali S.},
title = {Anomaly Detection Methods for Categorical Data: A Review},
year = {2019},
issue_date = {March 2020},
publisher = {Association for Computing Machinery},
address = {New York, NY, USA},
volume = {52},
number = {2},
issn = {0360-0300},
doi = {10.1145/3312739},
journal = {ACM Comput. Surv.},
month = may,
articleno = {38},
numpages = {35}
}

@inbook{SVD,
author = {Brunton, Steven and Kutz, J.},
year = {2019},
month = {02},
pages = {3-46},
title = {Singular Value Decomposition (SVD)},
isbn = {9781108422093},
doi = {10.1017/9781108380690.002},
publisher = {Cambridge University Press}
}

@book{box2015time,
  title={Time series analysis: forecasting and control},
  author={Box, George EP and Jenkins, Gwilym M and Reinsel, Gregory C and Ljung, Greta M},
  year={2015},
  publisher={John Wiley \& Sons}
}

@article{LJUNG20201175,
title = {Deep Learning and System Identification},
journal = {IFAC-PapersOnLine},
volume = {53},
number = {2},
pages = {1175-1181},
year = {2020},
note = {21st IFAC World Congress},
issn = {2405-8963},
doi = {10.1016/j.ifacol.2020.12.1329},
author = {Lennart Ljung and Carl Andersson and Koen Tiels and Thomas B. Schön},
}

@article{
doi:10.1073/pnas.1517384113,
author = {Steven L. Brunton  and Joshua L. Proctor  and J. Nathan Kutz },
title = {Discovering governing equations from data by sparse identification of nonlinear dynamical systems},
journal = {Proceedings of the National Academy of Sciences},
volume = {113},
number = {15},
pages = {3932-3937},
year = {2016},
doi = {10.1073/pnas.1517384113}}

@article{10.1093/genetics/162.4.2025,
    author = {Beaumont, Mark A and Zhang, Wenyang and Balding, David J},
    title = {Approximate Bayesian Computation in Population Genetics},
    journal = {Genetics},
    volume = {162},
    number = {4},
    pages = {2025-2035},
    year = {2002},
    month = {12},
    issn = {1943-2631},
    doi = {10.1093/genetics/162.4.2025},
}

@article{
doi:10.1073/pnas.1912789117,
author = {Kyle Cranmer  and Johann Brehmer  and Gilles Louppe },
title = {The frontier of simulation-based inference},
journal = {Proceedings of the National Academy of Sciences},
volume = {117},
number = {48},
pages = {30055-30062},
year = {2020},
doi = {10.1073/pnas.1912789117}}

@InProceedings{pmlr-v97-greenberg19a,
  title = 	 {Automatic Posterior Transformation for Likelihood-Free Inference},
  author =       {Greenberg, David and Nonnenmacher, Marcel and Macke, Jakob},
  booktitle = 	 {Proceedings of the 36th International Conference on Machine Learning},
  pages = 	 {2404--2414},
  year = 	 {2019},
  editor = 	 {Chaudhuri, Kamalika and Salakhutdinov, Ruslan},
  volume = 	 {97},
  series = 	 {Proceedings of Machine Learning Research},
  month = 	 {09--15 Jun},
  publisher =    {PMLR},
  url = 	 {https://proceedings.mlr.press/v97/greenberg19a.html}
}

@inproceedings{NIPS2016_6aca9700,
 author = {Papamakarios, George and Murray, Iain},
 booktitle = {Advances in Neural Information Processing Systems},
 editor = {D. Lee and M. Sugiyama and U. Luxburg and I. Guyon and R. Garnett},
 pages = {},
 publisher = {Curran Associates, Inc.},
 title = {Fast $\epsilon$ -free Inference of Simulation Models with Bayesian Conditional Density Estimation},
 volume = {29},
 year = {2016}
}

@article{10.1093/mnras/stz2006,
    author = {Humphries, J and Vazan, A and Bonavita, M and Helled, R and Nayakshin, S},
    title = {Constraining the initial planetary population in the gravitational instability model},
    journal = {Monthly Notices of the Royal Astronomical Society},
    volume = {488},
    number = {4},
    pages = {4873-4889},
    year = {2019},
    month = {07},
    issn = {0035-8711},
    doi = {10.1093/mnras/stz2006},
}

@article{Killick01122012,
author = {R. Killick and P. Fearnhead and I. A. Eckley},
title = {Optimal Detection of Changepoints With a Linear Computational Cost},
journal = {Journal of the American Statistical Association},
volume = {107},
number = {500},
pages = {1590--1598},
year = {2012},
publisher = {ASA Website},
doi = {10.1080/01621459.2012.737745},
}

@article{hong2024multi,
title = {Multi‐objective evolutionary optimization for hardware‐aware neural network pruning},
journal = {Fundamental Research},
volume = {4},
number = {4},
pages = {941-950},
year = {2024},
issn = {2667-3258},
doi = {10.1016/j.fmre.2022.07.013},
author = {Wenjing Hong and Guiying Li and Shengcai Liu and Peng Yang and Ke Tang}
}

@ARTICLE{chen2023multi,
  author={Chen, Wenjie and Hong, Wenjing and Zhang, Hu and Yang, Peng and Tang, Ke},
  journal={IEEE Transactions on Automation Science and Engineering}, 
  title={Multi-Fidelity Simulation Modeling for Discrete Event Simulation: An Optimization Perspective}, 
  year={2023},
  volume={20},
  number={2},
  pages={1156-1169},
  doi={10.1109/TASE.2022.3173296}}

@misc{zhong2025representation,
    title={Representation Learning of Limit Order Book: A Comprehensive Study and Benchmarking},
    author={Muyao Zhong and Yushi Lin and Peng Yang},
    year={2025},
    eprint={2505.02139},
    archivePrefix={arXiv},
    primaryClass={cs.CE}
}

@misc{huang2025learning,
    title={Learning Representations from Heterogeneous Data for Robust Heart Rate Modeling.},
    author={Zicheng Xie and Zhengdong Huang and Wentao Tian and Jingyu Liu and Lunhong Dong and and Peng Yang},
    year={2025},
    eprint={2508.21785},
    archivePrefix={arXiv},
    primaryClass={cs.LG}
}

@misc{ren2025from,
    title={From Linear to Hierarchical: Evolving Tree-structured Thoughts for Efficient Alpha Mining},
    author={Junji Ren and Junjie Zhao and Shengcai Liu and Peng Yang},
    year={2025},
    eprint={2508.16334},
    archivePrefix={arXiv},
    primaryClass={cs.CE}
}

@ARTICLE{zhao2025quantfactor,
  author={Zhao, Junjie and Zhang, Chengxi and Qin, Min and Yang, Peng},
  journal={IEEE Transactions on Signal Processing}, 
  title={QuantFactor REINFORCE: Mining Steady Formulaic Alpha Factors With Variance-Bounded REINFORCE}, 
  year={2025},
  volume={73},
  number={},
  pages={2448-2463},
  doi={10.1109/TSP.2025.3576781}}

@misc{zhao2025learning,
      title={Learning from Expert Factors: Trajectory-level Reward Shaping for Formulaic Alpha Mining}, 
      author={Junjie Zhao and Chengxi Zhang and Chenkai Wang and Peng Yang},
      year={2025},
      eprint={2507.20263},
      archivePrefix={arXiv},
      primaryClass={cs.LG},
      url={https://arxiv.org/abs/2507.20263}, 
}

@misc{lin2025detecting,
      title={Detecting Multilevel Manipulation from Limit Order Book via Cascaded Contrastive Representation Learning}, 
      author={Yushi Lin and Peng Yang},
      year={2025},
      eprint={2508.17086},
      archivePrefix={arXiv},
      primaryClass={q-fin.CP},
      url={https://arxiv.org/abs/2508.17086}, 
}

@Article{yang2024reducing,
author={Yang, Peng
and Zhang, Laoming
and Liu, Haifeng
and Li, Guiying},
title={Reducing idleness in financial cloud services via multi-objective evolutionary reinforcement learning based load balancer},
journal={Science China Information Sciences},
year={2024},
month={Jan},
day={25},
volume={67},
number={2},
pages={120102},
issn={1869-1919},
doi={10.1007/s11432-023-3895-3},
}

@inproceedings{qian2013analysis,
author = {Qian, Chao and Yu, Yang and Zhou, Zhi-Hua},
title = {An analysis on recombination in multi-objective evolutionary optimization},
year = {2011},
isbn = {9781450305570},
publisher = {Association for Computing Machinery},
address = {New York, NY, USA},
doi = {10.1145/2001576.2001852},
booktitle = {Proceedings of the 13th Annual Conference on Genetic and Evolutionary Computation},
pages = {2051–2058},
numpages = {8},
location = {Dublin, Ireland},
series = {GECCO '11}
}

@article{bian2025stochastic,
title = {Stochastic population update can provably be helpful in multi-objective evolutionary algorithms},
journal = {Artificial Intelligence},
volume = {341},
pages = {104308},
year = {2025},
issn = {0004-3702},
doi = {10.1016/j.artint.2025.104308},
author = {Chao Bian and Yawen Zhou and Miqing Li and Chao Qian}
}

@inproceedings{qian2015subset,
 author = {Qian, Chao and Yu, Yang and Zhou, Zhi-Hua},
 booktitle = {Advances in Neural Information Processing Systems},
 editor = {C. Cortes and N. Lawrence and D. Lee and M. Sugiyama and R. Garnett},
 pages = {},
 publisher = {Curran Associates, Inc.},
 title = {Subset Selection by Pareto Optimization},
 volume = {28},
 year = {2015}
}

@inproceedings{xue2022multi,
 author = {Xue, Ke and Xu, Jiacheng and Yuan, Lei and Li, Miqing and Qian, Chao and Zhang, Zongzhang and Yu, Yang},
 booktitle = {Advances in Neural Information Processing Systems},
 editor = {S. Koyejo and S. Mohamed and A. Agarwal and D. Belgrave and K. Cho and A. Oh},
 pages = {20147--20161},
 publisher = {Curran Associates, Inc.},
 title = {Multi-agent Dynamic Algorithm Configuration},
 volume = {35},
 year = {2022}
}

@book{zhou2019evolutionary,
  title={Evolutionary learning: Advances in theories and algorithms},
  author={Zhou, Zhi-Hua and Yu, Yang and Qian, Chao},
  year={2019},
  publisher={Springer}
}

@ARTICLE{li2025morphing,
  author={Li, Bingdong and Di, Zixiang and Yang, Yanting and Qian, Hong and Yang, Peng and Hao, Hao and Tang, Ke and Zhou, Aimin},
  journal={IEEE Transactions on Evolutionary Computation}, 
  title={It’s Morphing Time: Unleashing the Potential of Multiple LLMs via Multi-Objective Optimization}, 
  year={2025},
  volume={},
  number={},
  pages={1-1},
  doi={10.1109/TEVC.2025.3613937}}

@ARTICLE{li2025causal,
  author={Li, Bingdong and Yang, Yanting and Yang, Peng and Li, Guiying and Tang, Ke and Zhou, Aimin},
  journal={IEEE Transactions on Evolutionary Computation}, 
  title={Causal Inference-Based Large-Scale Multiobjective Optimization}, 
  year={2025},
  volume={29},
  number={2},
  pages={444-458},
  doi={10.1109/TEVC.2025.3529938}}

@misc{li2025surrogate,
      title={Surrogate-Assisted Evolutionary Reinforcement Learning Based on Autoencoder and Hyperbolic Neural Network}, 
      author={Bingdong Li and Mei Jiang and Hong Qian and Ke Tang and Aimin Zhou and Peng Yang},
      year={2025},
      eprint={2505.19423},
      archivePrefix={arXiv},
      primaryClass={cs.LG},
      url={https://arxiv.org/abs/2505.19423}, 
}

@misc{li2025simloblearningrepresentationslimited,
      title={SimLOB: Learning Representations of Limited Order Book for Financial Market Simulation}, 
      author={Yuanzhe Li and Yue Wu and Muyao Zhong and Shengcai Liu and Peng Yang},
      year={2025},
      eprint={2406.19396},
      archivePrefix={arXiv},
      primaryClass={cs.CE},
      url={https://arxiv.org/abs/2406.19396}, 
}

@ARTICLE{11073808,
  author={Wang, Chenkai and Ren, Junji and Yang, Peng},
  journal={IEEE Transactions on Computational Social Systems}, 
  title={Alleviating Nonidentifiability: A High-Fidelity Calibration Objective for Financial Market Simulation With Multivariate Time Series Data}, 
  year={2025},
  volume={12},
  number={6},
  pages={4910-4922},
  doi={10.1109/TCSS.2025.3574236}}

@ARTICLE{10969534,
  author={Yang, Peng and Ren, Junji and Wang, Feng and Tang, Ke},
  journal={Complex System Modeling and Simulation}, 
  title={Towards calibrating financial market simulators with high-frequency data}, 
  year={2025},
  volume={},
  number={},
  pages={1-16},
  doi={10.23919/CSMS.2025.0002}}

\end{document}